\documentclass{article}
\pdfoutput=1
\usepackage[utf8]{inputenc}
\usepackage[a4paper, margin=1in]{geometry}
\usepackage[active]{srcltx}
\usepackage{babel}
\usepackage{verbatim}
\usepackage{mathtools}
\usepackage{bm}
\usepackage{amsmath}
\usepackage{amssymb}
\usepackage{microtype,comment}
\usepackage{graphicx}
\usepackage{subfigure}
\usepackage{booktabs} 
\usepackage{hyperref}
\usepackage{verbatim}
\usepackage{mathtools}
\usepackage{bm}
\usepackage{amsmath}
\usepackage{amssymb} 
\usepackage{amsthm}
\usepackage{babel}
\usepackage{amsthm}
\usepackage{dsfont}
\usepackage{array}
\usepackage{mathrsfs}
\usepackage{color}
\usepackage{natbib}
\usepackage{enumerate}
 \bibpunct[, ]{(}{)}{,}{a}{}{,}%
 \def\BIBand{and}%
 
\theoremstyle{plain} 
\newtheorem{lemma}{\textbf{Lemma}} 

\newtheorem{theorem}{\textbf{Theorem}}
\newtheorem{corollary}{\textbf{Corollary}}

\theoremstyle{definition}
\theoremstyle{remark}

\title{Concentration of Contractive Stochastic Approximation and Reinforcement Learning}
\author{Siddharth Chandak, Vivek S.\ Borkar\footnote{Work supported in part by a S.\ S.\ Bhatnagar Fellowship from Council of Scientific and Industrial Research, Government of India}, Parth Dodhia\\Electrical Engineering, Indian Institute of Technology Bombay, Mumbai, India\\chandak1299@gmail.com, borkar.vs@gmail.com, parthdoh@gmail.com}

\date{March 2022}

\begin{document}
\maketitle
\begin{abstract}
    Using a martingale concentration inequality, concentration bounds `from time $n_0$ on' are derived for stochastic approximation algorithms with contractive maps and both martingale difference and Markov noises. These are applied to reinforcement learning algorithms, in particular to asynchronous Q-learning and TD(0).
\end{abstract}

asynchronous Q-learning; TD(0)

\section{Introduction}\label{sec-intro}

 In recent years, there has been a lot of interest in obtaining bounds for finite time behavior of reinforcement learning algorithms. These are either moment bounds, e.g., mean square error after finitely many samples, or high probability concentration bounds. A representative, but possibly non-exhaustive sample is: \citet{Bhandari, Chen, Chen2, Giannakis,Gugan2, Gugan3,  Li-QL1, Li-QL2, Mansour, Prashanth, Qu, Sidford, Srikant, Wain1, Wain2}.
A parallel activity in stochastic approximation theory (of which most reinforcement learning algorithms are special instances) seeks to get a concentration bound for the iterates \textit{from some time on}, to be precise, `\textit{for all $n \geq n_0$ for a suitably chosen $n_0$}' \citep{Borkar0, Gugan, Kamal}.  See \citet{BorkarQ} for an application to reinforcement learning. 

Inspired by \citet{Chen}, one of us considered stochastic approximation involving contractive maps and martingale noise, and derived such concentration bounds `from some $n_0$ on' for this class of algorithms \citep{Borkar-conc}. In addition, \citet{Borkar-conc}  indicated how to stitch such bounds with finite time bounds to get concentration bounds for all time. This covered in particular synchronous Q-learning for discounted cost and some related schemes. But it did not cover the asynchronous case, which is of greater importance. Nor could it cover some other algorithms such as TD(0). The missing link was the absence of the so called `Markov noise' in stochastic approximation, originally introduced in \citet{Meerkov} \textcolor{black}{(see also \citet{Kushner} for a landmark article and \cite{Benveniste} for a book length treatment).} The present work fills in this lacuna, extending the applicability of this program to a much larger class of algorithms. In fact, we work out in detail the cases of asynchronous Q-learning and TD(0).

\textcolor{black}{There is also a parallel body of work which seeks bound, either finite time or asymptotic (e.g., in terms of regret) on the difference between the value function under the learned policy and the optimal value function \citep{Chi_Jin, Yang, Lin_Yang}. Once again, this is distinct from our objective, which is to obtain a \textit{high probability} bound valid \textit{for all time from some time on}.}

 To clarify further,  conventional concentration or sample complexity bounds get bounds, e.g. mean square or `with high probability' bounds, on the error from the target after $n$ iterations starting from time zero. The asymptotic regret bounds get an upper bound or lower bound or both on how some measure of cumulative error grows with time in an asymptotic sense. Our bounds differ from both. They are `all time bounds' in the sense that they give a high probability bound for the iterates to remain in a prescribed small neighborhood of the target \textit{for all time from some time $n_0$ onwards}. In fact the requirement `from some $n_0$ onwards' is dictated by the fact that the decreasing stepsize needs to be sufficiently small from $n_0$ on. Thus if the stepsize is sufficiently small from the beginning, this qualification can be dropped. Alternatively, we can stitch our bounds with one of the existing finite time bounds to obtain all time bounds. We illustrate this possibility with an example in section \ref{sec-conseq}. The `time $n_0$ on' bound depends on the norm of the iterate at time $n_0$, but this in turn can be bounded in terms of the norm of the initialization. As for regret bounds, these are for cumulative error and are  typically `almost sure' or `in mean' and asymptotic in nature unlike our bounds, which are from some time on, but with high probability.

The rest of the paper is organized as follows. Section \ref{sec-main} sets up and states the main result, also highlighting an important special case.  The result is proved in Section \ref{sec-proof}. Section \ref{sec-conseq} presents some consequences of the theorem. In Section \ref{sec-apps}, we apply our main result to asynchronous Q-learning and the TD(0) algorithm.  A concluding section highlights some future directions. Finally, there are two appendices : Appendix A states a martingale concentration inequality used in our proof whereas Appendix B details a technical issue left out of the main text for ease of reading.

Throughout this work, $\|\cdot\|$ denotes any compatible norm on $\mathcal{R}^d$. We use $\theta$ to denote the zero vector in $\mathcal{R}^d$. The $\ell$\textsuperscript{th} component of a vector $x$ and a vector valued function $h(\cdot)$ are denoted by $x(\ell)$ and $h^{\ell}(\cdot)$ respectively. We use the convention $\frac{0}{0}=0$ throughout this work.

\section{Main Result}\label{sec-main}

We state and prove our main theorem in this section, after setting up the notation and assumptions. The assumptions are specifically geared for the reinforcement learning applications that follow in Section \ref{sec-apps}, as will become apparent. 

Consider the iteration 
\begin{equation}\label{iteration}
    x_{n+1}=x_n+a(n)\big(F(x_n,Y_n)-x_n+M_{n+1}(x_n)\big), n\geq 0,
\end{equation}
for $x_n=[x_n(1),...,x_n(d)]^T\in\mathcal{R}^d$. Here:
\begin{itemize}
    \item $\{Y_n\}$ is the `Markov noise' taking values in a finite state space $S$, i.e.,
    \begin{eqnarray*}
        P(Y_{n+1}|Y_m,x_m, m\leq n)&=&P(Y_{n+1}|Y_n,x_n)\\
        &=&p_{x_n}(Y_{n+1}|Y_n), \ n\geq 0,
    \end{eqnarray*}
where for each $w$, $p_w(\cdot|\cdot)$ is the transition probability of an irreducible Markov chain on $S$ with unique stationary distribution $\pi_w$. We assume that the map $w\mapsto p_w(j|i)$ is Lipschitz in the following sense:
$$\sum_{j\in S} |p_w(j|i)-p_v(j|i)|\leq L_1 \|w-v\|,\ \forall i\in S, w, v\in\mathcal{R}^d,$$
for some $L_1>0$. This implies that the map $w\mapsto \pi_w(i)$ is similarly Lipschitz, i.e.,
$$\sum_{i\in S}|\pi_w(i)-\pi_v(i)|\leq L_2 \|w-v\|,\ \forall i\in S, w, v\in\mathcal{R}^d,$$
for some $L_2>0$. See part (ii) of Appendix B for some bounds on $L_2$.

    \item $\{M_n(x)\}$ is, for each $x\in\mathcal{R}^d$, an $\mathcal{R}^d$-valued martingale difference sequence parametrized by $x$, with respect to the increasing family of $\sigma$-fields $\mathcal{F}_n\coloneqq\sigma(x_0, Y_m, M_m(x), x\in\mathcal{R}^d,m\leq n)$, $n\geq0$. That is,
    \begin{equation}\label{MDS}
        E[M_{n+1}(x)|\mathcal{F}_n]=\theta \: \ \mathrm{a.s.} \: \ \forall \; x,n,
    \end{equation}
    where $\theta$ denotes the zero vector. We also assume the componentwise bound:
    \begin{equation}\label{MDS_assumption}
        |M_n^\ell(x)|\leq K_0(1+\|x\|) \:\textrm{a.s.} \: \forall \; x,n,l,
    \end{equation}
    for some $K_0>0$.
    \item 
    $F(\cdot) = [F^1(\cdot), \cdots , F^d(\cdot)]^T:\mathcal{R}^d\times S\rightarrow\mathcal{R}^d$ satisfies
    \begin{equation}\label{Contraction}
        \|\sum_{i\in S}\pi_w(i)(F(x,i)-F(z,i))\|\leq \alpha \|x-z\|, \ \forall \ x,z,w\in\mathcal{R}^d,
    \end{equation}
    for some $\alpha\in(0,1)$. By the Banach contraction mapping theorem, (\ref{Contraction}) implies that $\sum_i\pi_w(i)F(\cdot,i)$ has a unique fixed point $x^*_w\in\mathcal{R}^d$ (i.e., $\sum_i\pi_w(i)F(x^*_w,i)=x^*_w)$. We assume that this fixed point is independent of $w$, i.e., there exists a $x^*$ such that
    \begin{equation}\label{fixed_point}
        \sum_{i\in S}\pi_w(i)F(x^*,i)=x^*, \ \forall \;w\in\mathcal{R}^d.
    \end{equation}
     We also assume that the map $x\mapsto F^\ell(x,i)$ is Lipschitz, uniformly in $i$ and $\ell$. Let the common Lipschitz constant be $L_3>0$, i.e., 
    $$|F^\ell(x,i)-F^\ell(z,i)|\leq L_3 \|x-z\|,\ \forall i\in S, \ell\in\{1,\cdots, d\}, x,z\in\mathcal{R}^d.$$
    Furthermore, $\widetilde{F}_n(x,Y_n)\coloneqq F(x,Y_n)+M_{n+1}(x)$ is assumed to satisfy
    \begin{equation}\label{F_tilde}
        \|\widetilde{F}_n(x,Y_n)\|\leq K+\alpha\|x\| \: \ \textrm{a.s.}.
    \end{equation}
    \item $\{a(n)\}$ is a sequence of non-negative stepsizes satisfying the conditions
    \textcolor{black}{\begin{equation}
     a(n) \to 0, \    \sum_na(n)=\infty,
    \end{equation}}
and is assumed to be eventually non-increasing, i.e., there exists $n^*\geq1$ such that $a(n+1)\leq a(n) \ \forall \; n\geq n^*$. Since $a(n)\to0$, there exists $n^\dagger$ such that $a(n)<1$ for all $n>n^\dagger$. Observe that we do not require the classical square-summability condition in stochastic approximation, viz., $\sum_na(n)^2 < \infty$. This is because the contractive nature of our iterates gives us an additional handle on errors by putting less weight on past errors. A similar effect was observed in \cite{Gugan}. 

We further assume that $a(n)=\Omega(1/n)$. So, $a(n)\geq \frac{d_1}{n}$ for all $n\geq n_1$ for some $n_1$ and $d_1>0$. We also assume that there exists $0<d_2 \leq1$ such that $a(n)=\mathcal{O}\left(\left(\frac{1}{n}\right)^{d_2}\right)$, i.e., $a(n)\leq d_3\left(\frac{1}{n}\right)^{d_2}$ for all $n\geq n_2$ for some $n_2$ and $d_3>0$. Larger values of $d_1$ and $d_2$ and smaller values of $d_3$ improve the main result presented below. The role this assumption plays in our bounds will become clear later. Define $N\coloneqq \max\{n^*, n^\dagger, n_1, n_2\}$, i.e., $a(N)<1$, $a(n)$ is non-increasing after $N$ and $\frac{d_1}{n}\leq a(n)\leq d_3\left(\frac{1}{n}\right)^{d_2}, \forall\ n\geq N$.
\end{itemize}

\ \ \ For $n\geq0$, we further define:
\begin{eqnarray*}
    b_k(n) &=& \sum_{m=k}^na(m), \ 0\leq k\leq n<\infty, \\
   \color{black}\beta_k(n) &\color{black}=& \color{black}\begin{cases}
      \frac{1}{k^{d_2-d_1}n^{d_1}}, & \text{if}\ d_1\leq d_2 \\
      \frac{1}{n^{d_2}}, & \text{otherwise},
    \end{cases} \\
    \kappa(d) &=& \|\textbf{1}\|, \ \textbf{1} := [1,1,\cdots , 1]^T \in \mathcal{R}^d, \ d \geq 1.
\end{eqnarray*}
Our main result is as follows: 

\begin{theorem}\label{Main_Theorem}
Let $n_0\geq N$. Then there exist finite positive constants $c_1$, $c_2$ and $D$, depending on $\|x_{N}\|$, such that for $\delta>0$, $n\geq n_0$ and $C=e^{\kappa(d)\left(K_0\left(1+\|x_{N}\|+\frac{K}{1-\alpha}\right)+c_2\right)}$,  
\begin{enumerate}[(a)]
    \item The inequality
\begin{equation}
    \|x_n-x^*\|\leq e^{-(1-\alpha)b_{n_0}(n-1)}\|x_{n_0}-x^*\|+\frac{\delta+a(n_0)c_1}{1-\alpha},
\end{equation}
holds with probability exceeding \begin{eqnarray}
&&1 \ - \  2d\sum_{m=n_0+1}^ne^{-D\delta^2/\beta_{n_0}(m)}, \ 0 < \delta \leq C, \label{probbound0} \\
&& 1 \ - \  2d\sum_{m=n_0+1}^ne^{-D\delta/\beta_{n_0}(m)}, \ \ \ \delta > C. \label{probbound}
\end{eqnarray}
\item In particular,
\begin{equation}
\|x_n - x^*\| \leq e^{-(1 - \alpha)b_{n_0}(n-1)}\|x_{n_0} - x^*\| + \frac{\delta+a(n_0)c_1}{1 - \alpha} \ \forall \; n \geq n_0,
\label{mainbound3}
\end{equation}
 with probability exceeding
\begin{eqnarray}
&&1 \ - \  2d\sum_{m\geq n_0+1}e^{-D\delta^2/\beta_{n_0}(m)}, \ 0 < \delta \leq C, \label{probbound1} \\
&& 1 \ - \  2d\sum_{m\geq n_0+1}e^{-D\delta/\beta_{n_0}(m)}, \ \ \ \delta > C. \label{probbound2}
\end{eqnarray}
\end{enumerate}
\end{theorem}

\textcolor{black}{For (\ref{probbound1})-(\ref{probbound2}), note that $\beta_{n_0}(m)$ is $O\left(\frac{1}{m^{c}}\right)$ for some constant $c$ and therefore $e^{-D/\beta_{n_0}(m)}$ is summable. Furthermore, $\sum_{m \geq n_0} e^{-D/\beta_{n_0}(m)} \downarrow 0$ as $n_0\uparrow\infty$.} 

\noindent \textbf{Remark 1:} An important special case of the above theorem is when $\{Y_n\}$ is a time homogeneous and uncontrolled Markov chain. In that case 
\begin{eqnarray*}
P(Y_{n+1}|Y_m,x_m, m\leq n)&=&P(Y_{n+1}|Y_n)\\
    &=&p(Y_{n+1}|Y_n), \ n\geq 0,
\end{eqnarray*}
with unique stationary distribution $\pi$.
Assumption (\ref{Contraction}) is now modified to
    \begin{equation}\label{special_contraction}
        \|\sum_{i\in S}\pi(i)(F(x,i)-F(z,i))\|\leq \alpha \|x-z\|, \ x,z\in\mathcal{R}^d
    \end{equation}
for some $\alpha\in(0,1)$. By the Banach contraction mapping theorem, (\ref{special_contraction}) implies that $\sum_i\pi(i)F(\cdot,i)$ has a unique fixed point $x^*\in\mathcal{R}^d$ (i.e., $\sum_i\pi(i)F(x^*,i)=x^*)$. Hence we no longer need assumption (\ref{fixed_point}).  The rest of the assumptions remain the same. The statement of the theorem also remains the same. 

\noindent \textbf{Remark 2:} The constants $c_1$ and $c_2$ depend on $\|x_N\|$  which in turn has a bound depending on $\|x_0\|$ that can be derived easily using the discrete Gronwall inequality under our assumptions. Also note that if $a(n)$ are decreasing and $a(0) < 1$, then we can take $N=0$. The calculations in Appendix B show that $c_1$ depends on quantities that essentially depend on the mean hitting time of a fixed state, which is also related to mixing. Thus one expects this constant to be lower for faster mixing chains. The exact dependence, however, is not simple.

\section{Proof of the Main Theorem}\label{sec-proof}

We begin with a lemma adapted from \citet{Borkar-conc} that bounds the iterates $\{x_n\}$ using (\ref{F_tilde}).
\begin{lemma}\label{Bound_x_n}
$\sup_{n\geq N}\|x_n\|\leq\|x_{N}\|+\frac{K}{1-\alpha},$ a.s.
\end{lemma}
\proof{Proof.} 
Using (\ref{F_tilde}), we have
\begin{eqnarray*}
\|x_{n+1}\|&=&\|(1-a(n))x_n+a(n)\widetilde{F}_n(x_n,Y_n)\| \nonumber\\
&\leq&(1-a(n))\|x_n\|+a(n)(K+\alpha\|x_n\|) \nonumber\\
&=&(1-(1-\alpha)a(n))\|x_n\|+a(n)K.
\end{eqnarray*}
For $n,m\geq N$, define $\psi(n,m)\coloneqq\prod_{k=m}^{n-1}(1-(1-\alpha)a(k))$ if $n>m$ and $1$ otherwise. Note that, since $a(N)<1$, $0\leq\psi(n,m)\leq1$ for all $n,m\geq N$. Then
\begin{equation*}
\|x_{n+1}\|-\frac{K}{1-\alpha}\leq (1-(1-\alpha)a(n))\left(\|x_n\|-\frac{K}{1-\alpha}\right).
\end{equation*}
Now $\|x_{N}\|\leq\|x_{N}\|+\frac{K}{1-\alpha}=\psi(N,N)\|x_{N}\|+\frac{K}{1-\alpha}$. Suppose
\begin{equation}\label{induction_step}
    \|x_n\|\leq\psi(n,N)\|x_{N}\|+\frac{K}{1-\alpha}
\end{equation}
for some $n\geq N$. Then,
\begin{eqnarray*}
\|x_{n+1}\|-\frac{K}{1-\alpha}&\leq&(1-(1-\alpha)a(n))\left(\psi(n,N)\|x_{N}\|+\frac{K}{1-\alpha}-\frac{K}{1-\alpha}\right)\nonumber\\
&\leq&\psi(n+1,N)\|x_{N}\|.
\end{eqnarray*}
By induction, (\ref{induction_step}) holds for all $n\geq N$, which completes the proof of Lemma \ref{Bound_x_n}. 
\endproof

\proof{Proof of Theorem \ref{Main_Theorem}.} 

Define $z_n$ for $n\geq n_0$ by
\begin{equation}
z_{n+1}=z_n+a(n)\left(\sum_{i\in \mathcal{S}}\pi_{x_n}(i)F(z_n,i)-z_n\right),
\end{equation}
where $z_{n_0}=x_{n_0}$. Then
\begin{eqnarray}
    x_{n+1}-z_{n+1}&=&(1-a(n))(x_n-z_n)  
    + a(n)M_{n+1} \nonumber\\
   & &+ \ a(n)\left(F(x_n,Y_n)-\sum_{i\in \mathcal{S}}\pi_{x_n}(i)F(z_n,i)\right) \label{step1} \\
    &=&(1-a(n))(x_n-z_n)  
    + \ a(n)M_{n+1} \nonumber\\
    &&+ \ a(n)\left(\sum_{i\in \mathcal{S}}\pi_{x_n}(i)(F(x_n,i)-F(z_n,i))\right)   \nonumber \\
    &&+ \ a(n)\left(F(x_n,Y_n)-\sum_{i\in \mathcal{S}}\pi_{x_n}(i)F(x_n,i)\right). \label{step2}
\end{eqnarray}
For $n,m\geq0$, let $\chi(n,m)=\prod_{k=m}^{n}(1-a(k))$ if $n\geq m$ and $1$ otherwise. For some $n\geq n_0$, we iterate the above for $n_0\leq m\leq n$ to obtain
\begin{eqnarray}\label{step3}
    x_{m+1}-z_{m+1}&=&\sum_{k=n_0}^m\chi(m,k+1)a(k)M_{k+1} \nonumber\\
    &&+\sum_{k=n_0}^m\chi(m,k+1)a(k)\left(\sum_{i\in \mathcal{S}}\pi_{x_k}(i)(F(x_k,i)-F(z_k,i))\right)\nonumber\\
    &&+\sum_{k=n_0}^m\chi(m,k+1)a(k)\left(F(x_k,Y_k)-\sum_{i\in \mathcal{S}}\pi_{x_k}(i)F(x_k,i)\right).
\end{eqnarray}

To simplify (\ref{step3}), we define $V(\cdot,\cdot) = [V^1(\cdot,\cdot) : V^2(\cdot,\cdot) : \cdots\cdots : V^d(\cdot, \cdot)]^T$ to be a solution of the Poisson equation:
\begin{equation}\label{Poisson_eqn}
    V(x,i)=F(x,i)-\sum_{j\in \mathcal{S}}\pi_x(j)F(x,j)+\sum_{j\in \mathcal{S}}p_x(j|i)V(x,j), \ i\in S.
\end{equation}
For $i_0\in S$, $\tau\coloneqq\min\{n>0:Y_n=i_0\}$ and $E_i[\cdot]=E[\cdot|Y_0=i]$, we know that 
\begin{equation}
V_1(x,i)=E_i\left[\sum_{m=0}^{\tau-1}(F(x,Y_m)-\sum_{j\in \mathcal{S}}\pi_x(j)F(x,j))\right], \ i\in S,
\end{equation}
is a solution to the Poisson equation (See, e.g., Lemma 4.2 and Theorem 4.2 of Section VI.4, pp.\ 85-91, of \citet{Borkar-Topics}). Thus $\|V_1(x,i)\|_{\infty}\leq2\max_i\|F(x,i)\|_{\infty}E_i[\tau]$. For an irreducible Markov chain with a finite state space, $E_i[\tau]$ is finite for all $i$ and hence the solution $V_1(x,i)$ is bounded for all $x,i$.  For each $x$ and $\ell$,  the Poisson equation specifies $V^\ell(x,\cdot)$ uniquely only up to an additive constant. Adding or subtracting a scalar $c^\ell(x)$ to $V^\ell(x,i)$ for each state $i$ still gives us a solution of the Poisson equation. Along with the additional constraint that $V(x,i_0)=0,\ \forall x$ and for a prescribed $i_0\in S$, the system of equations given by (\ref{Poisson_eqn}) has a unique solution. Henceforth $V$ refers to the unique solution of the Poisson equation with $V(x,i_0)=0, \ \forall x$. For this solution, the mapping $x\mapsto V^\ell(x,i)$ is Lipschitz for all $i\in S$, $\ell\in\{1,\cdots d\}$ and $x\in\mathcal{R}^d$ with the common constant $L$ (proof in Appendix B part (ii)).

We define $V_{max}$ as
\begin{equation}
    V_{max}\coloneqq\max_{\|x\|\leq\|x_{N}\|+\frac{K}{1-\alpha}}\max_{i\in S}\|V(x,i)\|.
\end{equation}
Similarly, we also define:
\begin{equation}
    V'_{max}\coloneqq\max_{\|x\|\leq\|x_{N}\|+\frac{K}{1-\alpha}}\max_{i\in S, 1\leq \ell\leq d}|V^\ell(x,i)|.
\end{equation}

Using the definition of $V$ to simplify the last term in (\ref{step3}), we have 
\begin{subequations}\label{V_expansion}
\begin{align}
    &\sum_{k=n_0}^m\chi(m,k+1)a(k)\left(F(x_k,Y_k)-\sum_i\pi_{x_k}(i)F(x_k,i)\right) \nonumber\\
    &\! =\sum_{k=n_0}^m\chi(m,k+1)a(k)\left(V(x_k,Y_k)-\sum_jp_{x_k}(j|Y_k)V(x_k,j)\right) \nonumber\\
    &\! =\sum_{k=n_0+1}^m\chi(m,k+1)a(k)\left(V(x_k,Y_k)-\sum_jp_{x_{k-1}}(j|Y_{k-1})V(x_k,j)\right) \label{V_martingale}\\
    &\; +\sum_{k=n_0+1}^m\left((\chi(m,k+1)a(k)-\chi(m,k)a(k-1))\sum_jp_{x_{k-1}}(j|Y_{k-1})V(x_k,j)\right) \label{V_2}\\
    &\; +\sum_{k=n_0+1}^m\chi(m,k)a(k-1)\left(\sum_jp_{x_{k-1}}(j|Y_{k-1})(V(x_k,j)-V(x_{k-1},j))\right)\label{V_3}\\
    &\; +\chi(m,n_0+1)a(n_0)V(x_{n_0},Y_{n_0})-\chi(m,m+1)a(m)\sum_jp_{x_m}(j|Y_m)V(x_m,j). \label{V_4}
\end{align}
\end{subequations}
Define $\widetilde{V}_k(x_k)\coloneqq V(x_k,Y_k)-\sum_jp_{x_{k-1}}(j|Y_{k-1})V(x_k,j)$ for $k\geq n_0$ and $0$ otherwise. This is a martingale difference sequence. We bound the norm of (\ref{V_2}) as follows:
\begin{align}\label{V_2_expansion}
&\left\|\sum_{k=n_0+1}^m\left((\chi(m,k+1)a(k)-\chi(m,k)a(k-1))\sum_{j\in S}p_{x_{k-1}}(j|Y_{k-1})V(x_k,j)\right)\right\|\nonumber\\
&\leq \left\|\sum_{k=n_0+1}^m((\chi(m,k+1)a(k)-\chi(m,k+1)a(k-1))\sum_{j\in S}p_{x_{k-1}}(j|Y_{k-1})V(x_k,j))\right\|\nonumber\\
&\;\;\;\;\;\;\;+ \left\|\sum_{k=n_0+1}^m((\chi(m,k+1)a(k-1)-\chi(m,k)a(k-1))\sum_{j\in S}p_{x_{k-1}}(j|Y_{k-1})V(x_k,j))\right\|\nonumber\\
&\leq \sum_{k=n_0+1}^m((a(k-1)-a(k))\chi(m,k+1)V_{max})+\sum_{k=n_0+1}^m((\chi(m,k+1)-\chi(m,k))a(k-1)V_{max})\nonumber\\
&\leq \sum_{k=n_0+1}^m((a(k-1)-a(k))V_{max})+\sum_{k=n_0+1}^m((\chi(m,k+1)-\chi(m,k))a(n_0)V_{max})\nonumber\\
&= (a(n_0)-a(m))V_{max} + (\chi(m,m+1)-\chi(m,n_0+1))a(n_0)V_{max}\nonumber\\
&\leq 2a(n_0)V_{max}.
\end{align}
The third inequality follows from $a(k-1)-a(k) \geq 0$  because  $a(k)$ is a non-increasing sequence for $k>n_0$, and $\chi(m,k+1)-\chi(m,k)$ is positive because $1\geq\chi(m,k+1)\geq\chi(m,k)$ for $m,k>n_0$, as $a(k)<1$ for $k>n_0$. 

We next obtain a bound on the norm of (\ref{V_3}). Using Lemma \ref{Bound_x_n}, we know that $\|x_n\| \leq \|x_{N}\|+\frac{K}{1-\alpha}, \ n \geq N$. For simplicity, define $K^*\coloneqq\|x_{N}\|+\frac{K}{1-\alpha}$. Note that $K^*$ is a \textit{random} constant due to its linear dependence on $\|x_{N}\|$.
Now,
\begin{eqnarray*}
\|x_k-x_{k-1}\|&=&\|a(k-1)(F(x_{k-1},Y_{k-1})-x_{k-1}+M_k(x_{k-1}))\| \nonumber \\
&\leq& a(k-1)(\|\widetilde{F}_{k-1}(x_{k-1},Y_{k-1})\|+\|x_{k-1}\|) \nonumber \\
&\leq& a(k-1)(K+\alpha\|x_{k-1}\|+\|x_{k-1}\|) \nonumber\\
&\leq& a(k-1)(K+(1+\alpha) K^*) \nonumber\\
&\leq& a(n_0)(K+(1+\alpha) K^*).
\end{eqnarray*}
For the last inequality, note that $k-1\geq n_0$ and hence, $a(k-1)\leq a(n_0).$ Now, note that for any $0<k\leq m$,
$$\chi(m,k)+\chi(m,k+1)a(k)=\chi(m,k+1),$$ and hence
$$\chi(m,n_0)+\sum_{k={n_0}}^m\chi(m,k+1)a(k)=\chi(m,m+1)=1.$$ This implies that 
\begin{equation}\label{Bound_chi_a}
    \sum_{k=n_0}^m\chi(m,k+1)a(k) \ \leq \ 1.
\end{equation}
\textcolor{black}{This implies that
\begin{align}
    &\left\|\sum_{k=n_0+1}^m\chi(m,k)a(k-1)\left(\sum_jp_{x_{k-1}}(j|Y_{k-1})(V(x_k,j)-V(x_{k-1},j))\right)\right\| \nonumber\\
    &\leq \sum_{k=n_0+1}^m\chi(m,k)a(k-1) \left\|\sum_jp_{x_{k-1}}(j|Y_{k-1})(V(x_k,j)-V(x_{k-1},j))\right\| \nonumber\\
    &\stackrel{(a)}{\leq} La(n_0)(K+(1+\alpha) K^*) \nonumber \\
    & \leq a(n_0)\left(L(K+2K^*)\right).
\end{align}
Inequality (a) is obtained using the Lipschitz nature of $V$.} Define constant $K^\dagger\coloneqq L(K+2K^*)$. Now, note that the norm of (\ref{V_4}) is directly bounded by $2a(n_0)V_{max}$. For simplicity, define $V_c(n_0)\coloneqq a(n_0)(4V_{max}+K^\dagger)$.

Recall that $\kappa(d)=\|\mathbf{1}\|$ where $\mathbf{1}$ is the d-vector of all 1's. Define
\begin{eqnarray*}
    \color{black}\Gamma_k & \color{black}= & \color{black} \kappa(d)\max_{l}\left|\sum_{r=n_0}^{k-1}\chi(k,r+1)a(r)(M^\ell_{r+1}(x_r)+\widetilde{V}^\ell_{r}(x_r))\right|, \\
    \textrm{and} \;\; \zeta_m &=& \max_{n_0\leq k\leq m} \Gamma_k.
\end{eqnarray*}
Returning to (\ref{step3}), we now have
\begin{eqnarray}
\|x_{m+1}-z_{m+1}\| 
& \color{black} \leq & \color{black} \left\|\sum_{k=n_0}^m\chi(m,k+1)a(k)(\sum_{i\in S}\pi_{x_k}(i)(F(x_k,i)-F(z_k,i)))\right\|+\Gamma_{m+1}+V_c(n_0) \nonumber\\
&\leq& \left\|\sum_{k=n_0}^m\chi(m,k+1)a(k)(\sum_{i\in S}\pi_{x_k}(i)(F(x_k,i)-F(z_k,i)))\right\|+\zeta_{m+1}+V_c(n_0) \nonumber\\
&\leq&\alpha\sum_{k=n_0}^m\chi(m,k+1)a(k)\|x_k-z_k\|+\zeta_{m+1}+V_c(n_0).
\end{eqnarray}
Let $x'_m=\sup_{n_0\leq k\leq m}\|x_k-z_k\|$ for $n_0\leq m< n$. Then using (\ref{Bound_chi_a}) and the fact that $\zeta_m\leq\zeta_n$,
\begin{eqnarray}\textbf{}
x'_{m+1}&\leq& \alpha x'_m\sum_{k=n_0}^{m}\chi(m,k+1)a(k)+\zeta_n+V_c(n_0)\nonumber\\
&\leq&\alpha x'_m+\zeta_n+V_c(n_0).
\end{eqnarray}
Since $x'_{m+1}\geq x'_m$ and $x'_{n_0}=\theta$, we have
\begin{equation}\label{Bound_X_m}
    x'_m\leq \frac{1}{1-\alpha}(\zeta_n+V_c(n_0)), \ n_0\leq m\leq n.
\end{equation}
By Lemma \ref{Bound_x_n}, $|M_{n+1}^\ell(x_n)|\leq K_0(1+\|x_{n_0}\|+\frac{K}{1-\alpha})$. Also,  $|\widetilde{V}^\ell_n(x_n)|\leq 2V'_{max}$. In Theorem \ref{thm-appendix} of the Appendix A, let
\begin{eqnarray*}
&C=e^{\kappa(d)(K_0(1+\|x_{n_0}\|+\frac{K}{1-\alpha})+2V'_{max})},\; \xi_{m,n}=\chi(n,m+1)a(m),\\
&\varepsilon=1,\; \gamma_1=1.
\end{eqnarray*}
\textcolor{black}{
Next, we choose suitable $\gamma_2$ and $\omega(n)$ such that
$$\max_{n_0\leq m\leq n}\xi_{m,n}=\max_{n_0\leq m\leq n}\chi(n,m+1)a(m) \leq \gamma_2\omega(n).$$
For this, we use our assumption that $\frac{d_1}{n}\leq a(n)\leq d_3\left(\frac{1}{n}\right)^{d_2}, \forall\ n\geq n_0$, to obtain:
\begin{align*}
    &\chi(n,m+1)=\prod_{k=m+1}^{n}(1-a(k))\leq \exp\left(-\sum_{k=m+1}^n a(k)\right)\leq \exp \left(-\sum_{k=m+1}^n \frac{d_1}{k}\right) \\
    &\;\;\;\;\;\;\;\;\;\;\;\;\;\;\;\;\;\;\;\;\;\;\;\;\;\;\leq \exp\left(-\int_{m+1}^n \frac{d_1}{y}dy\right)\leq \exp\left(d_1(\log(m+1)-\log(n))\right) \\
    &\;\;\;\;\;\;\;\;\;\;\;\;\;\;\;\;\;\;\;\;\;\;\;\;\;\; =\left(\frac{m+1}{n}\right)^{d_1}\\
    &\implies \max_{n_0\leq m\leq n} a(m)\chi(n,m+1)\leq \max_{n_0\leq m\leq n} d_3\left(\frac{1}{m}\right)^{d_2}\left(\frac{m+1}{n}\right)^{d_1} \leq \max_{n_0\leq m\leq n}d_3\left(\frac{1}{m}\right)^{d_2}\left(\frac{2m}{n}\right)^{d_1}.
\end{align*}
From the last inequality, $\gamma_2=d_32^{d_1}$ and $\omega(n)=\beta_{n_0}(n)$ satisfy the required conditions.
}

Then for $n_0<m\leq n$, a suitable constant $D>0$ and $\delta\in(0,C\gamma_1]$, we have
\textcolor{black}{
\begin{equation*}
P(\Gamma_m\geq\delta)\leq 2de^{-D\delta^2/\beta_{n_0}(m)},
\end{equation*}
and for $\delta>C\gamma_1$,
\begin{equation*}
P(\Gamma_m\geq\delta)\leq 2de^{-D\delta/\beta_{n_0}(m)}.
\end{equation*}}
\noindent The factor $d$ comes from the union bound along with Theorem \ref{thm-appendix}. \textcolor{black}{Applying union bound again, for $\delta\in(0,C\gamma_1]$, we now have 
\begin{equation}\label{Bound_zeta1}
P(\zeta_n\geq\delta)\leq 2d\sum_{m=n_0+1}^{n}e^{-D\delta^2/\beta_{n_0}(m)},
\end{equation}
and for $\delta>C\gamma_1$,
\begin{equation}\label{Bound_zeta2}
P(\zeta_n\geq\delta)\leq 2d\sum_{m=n_0+1}^{n}e^{-D\delta/\beta_{n_0}(m)}.
\end{equation}}

Since $\sum_i\pi_{x_n}(i)F(x^*,i)=x^*$,
\begin{equation*}
    z_{n+1}-x^*=(1-a(n))(z_n-x^*)+a(n)\sum_{i\in S}\pi_{x_n}(i)(F(z_n,i)-F(x^*,i)),
\end{equation*}
which implies
\begin{eqnarray*}
   \|z_{n+1}-x^*\|&\leq&(1-a(n))\|z_n-x^*\|+a(n)\|\sum_{i\in S}\pi_{x_n}(i)(F(z_n,i)-F(x^*,i))\|\\
   &\leq&(1-(1-\alpha)a(n))\|z_n-x^*\|.
\end{eqnarray*}
We then have
\begin{equation}\label{Bound_z_x*}
    \|z_n-x^*\|\leq \psi(n,n_0)\|x_{n_0}-x^*\|\leq e^{-(1-\alpha)b_{n_0}(n-1)}\|x_{n_0}-x^*\|.
\end{equation}
Using (\ref{Bound_X_m}) and the fact that $\|x_n-z_n\|\leq x'_n$, we have
\begin{equation}
    \|x_n-z_n\|\leq\frac{1}{1-\alpha}(\zeta_n+V_c(n_0)).
\end{equation}
This inequality along with (\ref{Bound_z_x*}) and the fact that $\zeta_n<\delta$ holds with probabilities given by (\ref{Bound_zeta1}) and (\ref{Bound_zeta2}), completes the proof \textcolor{black}{of part (a)} of Theorem \ref{Main_Theorem} with constants defined as $c_1=4V_{max}+K^\dagger$ and $c_2=2V'_{max}$. 

\textcolor{black}{
For part (b) of Theorem \ref{Main_Theorem}, to get bounds for all $n\geq n_0$, note that $P\left(\cap_{m=n_0+1}^{\infty} \{\zeta_m < \delta \}\right) = P\left(\cap_{m=n_0+1}^{\infty} \{\Gamma_m < \delta \}\right)$. Similar to the proof of part (a), applying union bound gives us the desired result. 
}
\endproof 

\section{Some Consequences}\label{sec-conseq}

In this section we briefly highlight some consequences of the foregoing as in \citet{Borkar-conc}. We first show that Theorem \ref{Main_Theorem} implies in particular the almost sure convergence of the iterates to $x^*$.
\begin{corollary}
$x_n\to x^* \; a.s.$
\end{corollary}
\proof{Proof.}

Let $\hat{b}(k)$ and $\delta(k)$ be two sequences such that
\[
0<\hat{b}(k)\downarrow 0 \quad \textrm{and} \quad 0<\delta(k)<\frac{(1-\alpha)\hat{b}(k)}{4},\; k\geq 1.
\]
Also, let $C(n,n_0,\delta)$ denote the expression on the right hand side of (\ref{mainbound3}). Then for each $\delta=\delta(k)$ in (\ref{mainbound3}), increase $n_0 = n_0(k)>0$ sufficiently so that  $a(n_0(k))c_1 < \frac{(1-\alpha)\hat{b}(k)}{4}$ and furthermore, (\ref{probbound1}), resp., (\ref{probbound2}) exceed $1-\frac{1}{k^2}$. Then pick $n_1(k)>n_0(k)$ such that
\begin{equation*}
    e^{-(1 - \alpha)b_{n_0(k)}(n_1(k)-1)}\|x_{n_0(k)} - x^*\| \leq \frac{\hat{b}(k)}{2}.
\end{equation*}
Thus $C(n_1(k), n_0(k), \delta(k)) < \hat{b}(k)$. This leads to
$$\sum_kP\left(\sup_{m\geq n_1(k)}\|x_m - x^*\| > \hat{b}(k)\right) \ \leq \ \sum_k\frac{1}{k^2} \ < \ \infty.$$
By the Borel-Cantelli lemma, $\|x_{n_1(k)} - x^*\| \leq \hat{b}(k)$ for $k$ sufficiently large, a.s. Since $\hat{b}(k) \downarrow 0$, it follows that $x_k \to x^*$ a.s.  
\endproof

The proof also shows that $\{b(k)\}$ serves as a `regret bound' for the `cost' $\|x_n - x^*\|$, though possibly not the tightest possible.
As indicated in the introduction,  the foregoing can be combined with existing finite time sample complexity bounds to obtain a concentration claim for all time.  The combined estimate  then yields a bound on how many iterates are needed in order to remain within a prescribed neighborhood of the target $x^*$ from some time on, with probability exceeding a prescribed lower bound. Suppose one has a finite time bound of the type (see, e.g., \citet{Chen}) 
$$E\left[\|x_n - x^*\|^2\right]\leq \upsilon(n) = o(n)$$
for a suitable $\upsilon(n)$. Let $\breve{K} = 2K$ and $1 > \nu > 0$. Pick $\delta > 0$ small enough, followed by $n_0$ large enough and then followed by $n_1 \geq n_0$, so that for $n \geq n_0$,
\begin{eqnarray}
P\left(\|x_{n} - x^*\| > \check{K}\right) &\leq& \frac{E\left[\|x_{n} - x^*\|^2\right]}{\breve{K}^2} \nonumber \\
&\leq& \frac{\upsilon(n_0)}{\breve{K}^2} \ < \ \frac{\nu}{2}, \label{Chen}
\end{eqnarray}
and, in addition, $(i)$ the r.h.s.\ of (\ref{probbound1})/(\ref{probbound2}) exceeds $1 - \frac{\nu}{2}$ and, $(ii)$ for $n \geq n_1$, the r.h.s.\ of (\ref{mainbound3}) does not exceed $\epsilon$.
Then, using our theorem above,
\begin{equation}P\left(\sup_{n \geq n_1} \|x_n - x^*\| \geq \epsilon\right) \leq \nu. \label{ultimate}\end{equation}
We shall exploit this simple fact in order to stitch our bound for $n \geq n_0$ with that of \citet{Chen} for $n = n_0$, which allows us to bound the $\nu$ on the right hand side. This allows us to estimate the number of samples $n_1$ required in order to ensure that the iterates remain in the $\epsilon$-neighborhood of $x^*$ thereafter, with probability $\geq 1 - \nu$. 

\textcolor{black}{
 We shall make this more precise for the special choice of $a(n) = \frac{b}{n+1}, n \geq 0$, with $b > 0$. The derivation is adapted from \cite{Borkar-conc}, included here for sake of completeness. 
From \cite{Chen}, $\upsilon(n)$ above is of the form
\begin{equation*}
\upsilon(n) = c_1\prod_{m=0}^{n-1}\left(1 - \frac{c_3b}{m+1}\right) + c_2\sum_{k=0}^{n-1}\prod_{m=k+1}^{n-1}\left(1 - \frac{c_3b}{m+1}\right)\frac{1}{(k+1)^2}
\end{equation*}
for suitable constants $c_1, c_2, c_3 > 0, c_2 < 1,$  given explicitly in \cite{Chen}. We assume that $b > 0$ satisfies the condition $b \leq c_2/c_3$ required in Theorem 2.1 of \cite{Chen}.  Let $\Gamma := c_3b$ which, because $c_2 < 1$ in \cite{Chen}, leads to $\Gamma < 1$. Using the facts $1 - x \leq e^{-x}$ and $\sum_{m=0}^{n-1}\frac{1}{m+1} = \Theta(\log n)$, and using $c$ to denote a generic constant that can change from place to place, we have
\begin{eqnarray*}
\upsilon(n) &\leq& c\Bigg(e^{-\Gamma\sum_{m=0}^{n-1}\frac{1}{m+1}}+\sum_{k=0}^{n-1}e^{-\Gamma\sum_{m=k+1}^{n-1}\frac{1}{m+1}}\frac{1}{(k+1)^2}\Bigg) \\
&\leq& \frac{c}{n^{\Gamma}}\left(1 + \sum_{k=0}^{n-1}\frac{1}{(k+1)^{2 - \Gamma}}\right) \ \leq \ \frac{c}{n^{\Gamma}} \ .
\end{eqnarray*}
With $c$ as in the r.h.s.\ above, let $n_0$ satisfy
\begin{equation}
\frac{c}{(n_0)^{\Gamma}} \leq \frac{\nu\breve{K}^2}{2}, \ \mbox{i.e.,} \ n_0 \geq \left(\frac{2c}{\nu\breve{K}^2}\right)^{\frac{1}{\Gamma}}. \label{n0one}
\end{equation}
Then (\ref{Chen}) holds.
Consider the specific choice of $a(n)=b/(n+1)$, $\beta_{n_0}(n) \leq \frac{1}{n_{0}^{1-b/2}n^{b/2}}$ when $1\geq b/2$ and $\beta_{n_0}(n) \leq 1/n$ when $1<b/2$, where $n\geq n_0$. Then the r.h.s.\ of (\ref{probbound1}) exceeds  $1 -  \frac{\nu}{2}$ if\footnote{With $\delta \in (0, C\varphi(0)]$. An analogous result holds for $\delta > C\varphi(0)$ with $\delta$ replacing $\delta^2$ in (\ref{n0three}).} 
\begin{equation}
2d\sum_{n\geq n_0}e^{-D\delta^2/\beta_{n_0}(n)} < \frac{\nu}{2}. \label{n0three}
\end{equation}
Choosing $n_0$ as in (\ref{n0one}), (\ref{n0three}), the bound (\ref{ultimate}) follows. }Similar approach can be used for combining our bound with other finite time bounds in order to obtain an `all time' bound. Recall also that if $\{a(n)\}$ are monotone and sufficiently small (i.e., $N=0$), our bound already holds for all time if $n_0=0$.

\section{Applications to Reinforcement Learning}\label{sec-apps}

In this section we apply the general results above for two important reinforcement learning algorithms, viz., asynchronous Q-learning and TD(0), and indicate some  related algorithms where they apply as well. In particular, these examples cannot be covered by the results of \citet{Borkar-conc} which does not cover Markov noise.

\subsection{Asynchronous Q-Learning}
We first apply the above theorem to asynchronous Q-learning \citep{Watkins, Dayan}. Consider a controlled Markov chain $\{X_n\}$ on a finite state space $S_1$, $|S_1|=s$, controlled by a control process $\{Z_n\}$  in a finite action space $A$, $|A|=r$. The controlled transition probability function $(i,j,u) \in S_1^2\times A \mapsto p(j|i,u) \in [0,1]$ satisfies $\sum_jp(j | i,u) = 1 \ \forall \; i,u.$
Thus
$$P(X_{n+1} = j | X_m, Z_m, m \leq n) = p(j | X_n, Z_n), \ a.s.$$
The objective is to minimize the discounted cost
$$E\left[\sum_{m=0}^\infty\gamma^mk(X_m, Z_m)\right],$$
where $\gamma \in (0,1)$ is the discount factor and $k: S\times U \mapsto \mathcal{R}$ is a prescribed `running cost' function. Let $a(n) > 0$ be as above. Then the Q-learning algorithm is:
\begin{equation}\label{QL}
    Q_{n+1}(i,u)=Q_n(i,u)+a(n)I\{X_n=i,Z_n=u\}\Big(k(i,u)+\gamma\min_aQ_n(X_{n+1},a)-Q_n(i,u)\Big)
\end{equation}
with arbitrary $Q_0(\cdot,\cdot)\geq0$. Note that $I\{X_n=i,Z_n=u\}$ is the indicator function defined as follows:
\begin{eqnarray*}
I\{X_n = i, Z_n = u\} &:=& 1 \ \ \  \mbox{if} \ X_n = i, \ Z_n = u, \\
&:=& 0 \ \ \  \mbox{otherwise}, \ \ n \geq 0.
\end{eqnarray*}
For application of our theorem, $(X_n,Z_n)$ together forms the Markov chain with the state space as $S_1\times A$ and the transition probabilities given by
\begin{eqnarray*}
    P(X_{n+1}=i,Z_{n+1}=u|X_n,Z_n)&=&p_{Q_n}(i,u|X_n,Z_n)\\
    &=&p(i|X_n,Z_n)\Phi_{Q_n}(u|X_{n+1}).
\end{eqnarray*}
\textcolor{black}{Here $p(\cdot|\cdot,\cdot)$ is as above and $\Phi_{Q_n}(\cdot|\cdot)$ is the randomized policy.  We make the additional assumption that the graph of the Markov chain remains irreducible under all control choices. We also assume that the map $Q\mapsto \Phi_Q(\cdot|\cdot)$ is Lipschitz and that $\Phi_Q(u|i)>0$ for all $u\in A, i\in S_1$. In the case of offline Q-learning, where the policy is fixed, $\Phi_{Q}$ is independent of $Q$ and automatically satisfies this assumption. Softmax Q-learning is an example of an online learning algorithm which satisfies the assumption \citep{Satinder}.} For a given $Q$, the stationary distribution $\pi_Q$ is $\pi_{Q}(i,u)=\pi_{\Phi_{Q}}(i)\Phi_{Q}(u|i)$ where $\pi_{\Phi_{Q}}$ is the stationary distribution of states corresponding to the policy $\Phi_{Q}$. 

We first rearrange the iteration in (\ref{QL}) to get it in the form of (\ref{iteration}) and then verify the assumptions. The iteration (\ref{QL}) can be rewritten as: 
\begin{eqnarray}
    Q_{n+1}(i,u)&=&Q_n(i,u)+a(n)I\{X_n=i,Z_n=u\}\Big(k(i,u)+\gamma\min_aQ_n(X_{n+1},a)-Q_n(i,u)\Big)\nonumber\\
    &=&Q_n(i,u)+a(n)I\{X_n=i,Z_n=u\}\times\nonumber\\
    &&\Bigg(k(i,u)+\gamma\sum_jp(j|i,u)\min_aQ_n(j,a)-Q_n(i,u)\nonumber\\
    &&\;\;\;\;\;+\gamma \left(\min_aQ_n(X_{n+1},a)-\sum_jp(j|i,u)\min_aQ_n(j,a)\right)\Bigg)\nonumber\\
    &=&Q_n(i,u)+a(n)\Big(F^{(i,u)}(Q_n,X_n,Z_n)-Q_n(i,u)+M_{n+1}^{(i,u)}(Q_n)\Big).
\end{eqnarray}
where $$F^{i,u}(Q,X,Z)=I\{X=i,Z=u\}\Big(k(i,u)+\gamma\sum_jp(j|i,u)\min_aQ(j,a)-Q(i,u)\Big)+Q(i,u)$$ and $$M_{n+1}^{i,u}(Q)=\gamma I\{X_n=i,Z_n=u\} (\min_aQ(X_{n+1},a)-\sum_jp(j|i,u)\min_aQ(j,a))\Big).$$ \textcolor{black}{We assume that $Q_0(\cdot,\cdot),k(\cdot,\cdot)\geq0$, and $a(n)\leq 1,\ \forall n$ which implies $Q_n(\cdot,\cdot)\geq 0$ for all $n$. Note that we make these assumptions for sake of simplicity and they can be dropped.} Define the family of $\sigma$-fields  $\mathcal{F}_n, n\geq0,$ by 
$$\mathcal{F}_n\coloneqq\sigma(Q_0,X_m,Z_m, m\leq n).$$ 
Then $\{M_n(Q),\mathcal{F}_n\}$ is a martingale difference sequence for each $Q$ satisfying (\ref{MDS_assumption}) for $K_0=1$, as is $\{M_n(Q_n),\mathcal{F}_n\}$. 

For ease of notation, we define $g(Q)=[g^{i,u}(Q)]$ where $$g^{i,u}(Q)\coloneqq k(i,u)+\gamma\sum_jp(j|i,u)\min_aQ(j,a).$$ Note that $g(Q)$ is a contraction in the maximum norm with $$\|g(Q)-g(Q')\|_\infty\leq\gamma\|Q-Q'\|_\infty.$$ We also define the diagonal matrix $\Lambda_Q$ with values $\pi_Q(i,u)$, i.e., the stationary probabilities of $(X_n,Z_n)$ corresponding to the policy chosen based on $Q$. Then for any $Q,Q_1,Q_2\in\mathcal{R}^{s\times r}$:
\begin{align}
    &\|\sum_{i\in S_1,u\in A}\pi_Q(i,u)\big(F(Q_1,i,u)-F(Q_2,i,u)\big)\|_\infty\nonumber\\ &=\|(I-\Lambda_Q)(Q_1-Q_2)+\Lambda_Q(g(Q_1)-g(Q_2))\|_\infty\nonumber\\
    &=\max_{i,u}\mid(1-\pi_Q(i,u))(Q_1(i,u)-Q_2(i,u))+\pi_Q(i,u)(g^{i,u}(Q_1)-g^{i,u}(Q_2))\mid\nonumber\\
    &\leq\max_{i,u}\Big((1-\pi_Q(i,u))\mid Q_1(i,u)-Q_2(i,u)\mid+\pi_Q(i,u)\mid g^{i,u}(Q_1)-g^{i,u}(Q_2)\mid\Big)\nonumber\\
    &\leq\max_{i,u}\Big((1-\pi_Q(i,u))\max_{j,v}\mid Q_1(j,v)-Q_2(j,v)\mid+\pi_Q(i,u)\gamma\max_{j,v}\mid Q_1(j,v)-Q_2(j,v)\mid\Big)\nonumber\\
    &\leq(1-(1-\gamma)\pi_{min})\|Q_1-Q_2\|_\infty,
\end{align}
where $\pi_{min}=\min_{Q\in\mathcal{R}^{sr}}\min_{i,u}\pi_Q(i,u)$. Since the stationary distribution $\pi_x$ is uniquely specified by a linear system continuously parametrized by $x$, it is continuous in $x$. Also, the facts that $\pi_x(i) > 0 \ \forall \ x,i,$ and $x, i$ take values in compact sets (by Lemma \ref{Bound_x_n}), together imply that $\pi_{min} := \min_{i,x}\pi_x(i) > 0$. So, $F$ satisfies assumption (\ref{Contraction}) with $\alpha=1-(1-\gamma)\pi_{min}\in(0,1)$, as $\pi_{min}\in(0,1)$. The fixed point of $\sum_{i,u}\pi(i,u)F(\cdot,i,u)$ is the vector $Q^*$ of the true Q-values and satisfies
\begin{equation*}
    Q^*(i,u)=k(i,u)+\gamma\sum_{j}p(j|i,u)\min_aQ^*(j,a).
\end{equation*}
This implies that $g(Q^*)=Q^*$. Furthermore, for any $Q\in\mathcal{R}^{sr}$,
\begin{eqnarray*}
\sum_{i\in S_1, u\in A}\pi_Q(i,u)F(Q^*,i,u)&=&(I-\Lambda_Q)Q^*+\Lambda_Q g(Q^*) \\
&=&(I-\Lambda_Q)Q^*+\Lambda_Q Q^*\\
&=&Q^*.
\end{eqnarray*}
Hence assumption (\ref{fixed_point}) is also satisfied. The map $Q\mapsto F(Q,\cdot)$ is also clearly Lipschitz. Also note that $\|Q^*\|_\infty\leq\|k\|_\infty/(1-\gamma)$. For simplicity, we assume that this bound holds for $Q_0$ as well and hence holds for $Q_n, \forall \; n\geq 0$, by induction. Thus assumption (\ref{F_tilde}) also holds with \textcolor{black}{$K=\left(\frac{\|k\|_\infty}{1-\gamma}\right)$}. Then Theorem \ref{Main_Theorem} gives us:

\begin{corollary}
Let $n_0\geq N$. Then there exist finite positive constants $c_1$, $c_2$ and $D$, depending on $\|Q_{N}\|$, such that for $\delta>0$ and $n\geq n_0$,  
\begin{enumerate}[(a)]
    \item The inequality 
    \begin{equation}
    \|Q_n-Q^*\|\leq e^{-(1-\alpha)b_{n_0}(n-1)}\|Q_{n_0}-Q^*\|+\frac{\delta+a(n_0)c_1}{1-\alpha}
\end{equation}
holds with probability exceeding \begin{eqnarray}
&&1 \ - \  2rs\sum_{m=n_0+1}^ne^{-D\delta^2/\beta_{n_0}(m)}, \ 0 < \delta \leq C, \label{Q_probbound0} \\
&& 1 \ - \  2rs\sum_{m=n_0+1}^ne^{-D\delta/\beta_{n_0}(m)}, \ \ \ \delta > C, \label{Q_probbound}
\end{eqnarray}
where $C=e^{\left(2\left(1+\|Q_{N}\|_\infty+\frac{\|k\|_\infty}{1-\alpha}\right)+c_2\right)}$.
    \item In particular,
\begin{equation}
\|Q_n - Q^*\| \leq e^{-(1 - \alpha)b_{n_0}(n-1)}\|Q_{n_0} - Q^*\| + \frac{\delta+a(n_0)c_1}{1 - \alpha} \ \forall n \geq n_0,
\label{Q_mainbound3}
\end{equation}
 with probability exceeding
\begin{eqnarray}
&&1 \ - \  2rs\sum_{m\geq n_0+1}e^{-D\delta^2/\beta_{n_0}(m)}, \ 0 < \delta \leq C, \label{Q_probbound1} \\
&& 1 \ - \  2rs\sum_{m\geq n_0+1}e^{-D\delta/\beta_{n_0}(m)}, \ \ \ \delta > C. \label{Q_probbound2}
\end{eqnarray}
\end{enumerate}

\end{corollary}

\subsection{TD(0)}
We next apply  Theorem 1 to the popular algorithm TD(0) for policy evaluation \citep{vanRoyTsitsiklis}. We fix a stationary policy  a priori and thus work with an uncontrolled Markov chain $\{Y_n\}$ on state space $S$ with transition probabilities $p(\cdot|\cdot)$ (the dependence on the policy is suppressed). Assume that the chain is irreducible with the stationary distribution $\pi=[\pi(1),...,\pi(s)], s=|S|$ and let $D\coloneqq$ the $s\times s$ diagonal matrix whose $i$\textsuperscript{th} diagonal entry is $\pi(i)$. The dynamic programming equation is 
\[
\Upsilon(i)=k(i)+\gamma\sum_jp(j|i)\Upsilon(j), \ i\in S,
\] which can be written as the following vector equation
\[
\Upsilon=k+\gamma P\Upsilon
\]
for $k=[k(1),...,k(s)]^T$ and $p=[[p(j|i)]]_{i,j\in S}\in \mathcal{R}^{s\times s}$.

$\Upsilon$ is approximated using a linear combination of linearly independent basis functions (feature vectors) $\phi_i:S\mapsto \mathcal{R}, 1\leq i\leq M$, with $s>>M\geq 1$. Thus $\Upsilon(i)\approx\sum_{m=1}^Mr(m)\phi_m(i)$, i.e., $\Upsilon\approx\Phi r$ where $r=[r(1),...,r(M)]^T$ and $\Phi$ is an $s\times M$ matrix whose $i$\textsuperscript{th} column is $\phi_i$. Since $\{\phi_i\}$ are linearly independent, $\Phi$ is full rank. Substituting this approximation into the dynamic programming
equation above leads to
\[
\Phi r\approx k+\gamma P\Phi r.
\]
But the RHS may not belong to the range of $\Phi$. So we use the following fixed point equation:
\begin{equation}\label{TD0-H-fixed}
    \Phi r=\Pi(k+\gamma P\Phi r)\coloneqq H(\Phi r),
\end{equation}
where $\Pi$ denotes the projection to Range($\Phi$) with respect to a suitable norm. Here we take projection with respect to the weighted norm 
$\|x\|_D\coloneqq(\sum_i\pi(i)|x(i)|^2)^{1/2}$
whereby the projection map is
\begin{equation}
    \Pi x\coloneqq\Phi(\Phi^TD\Phi)^{-1}\Phi^TDx.
\end{equation}
The invertibility of $\Phi^TD\Phi$ is guaranteed by the fact that $\Phi$ is full rank. Also, $\|Px\|_D\leq\|x\|_D$ (by Jensen's inequality) and $\|\Pi x\|_D\leq \|x\|_D$ (because $\Pi$ is a $\|\cdot\|_D$-projection).

The TD(0) algorithm is given by the recursion
\begin{equation}\label{TD0}
    r_{n+1}=r_n+a(n)\varphi(Y_n)\Big(k(Y_n)+\gamma\varphi(Y_{n+1})^Tr_n-\varphi(Y_n)^Tr_n\Big).
\end{equation}
Here $\varphi(i)^T\in\mathcal{R}^M$ for $i\in S$ denotes the $i$\textsuperscript{th} row of $\Phi$. We will apply our theorem to the iterates $r_n$ using the Euclidean norm (i.e., $\|r\|_2=(\sum_i r(i)^2)^{1/2}$).

Before moving forward, we make an assumption on $\Phi$ which is not restrictive as we argue later. Define $\Psi\coloneqq\Phi^T\sqrt{D}$ and let $\lambda_M$ be the largest singular value of $\Psi$, i.e., the largest eigenvalue of $\Psi\Psi^T$ and equivalently, of $\Psi^T\Psi$. Assume that
\begin{equation}\label{TD0_assumption}
    \lambda_M<\frac{\sqrt{2(1-\gamma)}}{(1+\gamma)}.
\end{equation}
Since the feature vectors can be scaled without affecting the algorithm (the weights $r(i)$ get scaled accordingly), this assumption does not restrict the algorithm. 

Rearrange iteration (\ref{TD0}) as
\begin{eqnarray}\label{TD0-modified}
r_{n+1}&=&r_n+a(n)\varphi(Y_n)\Big(k(Y_n)+\gamma\varphi(Y_{n+1})^Tr_n-\varphi(Y_n)^Tr_n\Big)\nonumber\\
&=&r_n+a(n)\bigg(\varphi(Y_n)k(Y_n)+\gamma\varphi(Y_n)\varphi(Y_{n+1})^Tr_n-\varphi(Y_n)\varphi(Y_n)^Tr_n+r_n-r_n\bigg)\nonumber\\
&=&r_n+a(n)\bigg(\varphi(Y_n)k(Y_n)+\gamma\varphi(Y_n)\sum_jp(j|Y_n)\varphi(j)^Tr_n-\varphi(Y_n)\varphi(Y_n)^Tr_n+r_n-r_n\nonumber\\
&&\;\;\;\;\;\;\;\;\;\;\;\;\;\;\;\;\;\;\;\;\;\;+\gamma\varphi(Y_n)\varphi(Y_{n+1})^Tr_n-\gamma\varphi(Y_n)\sum_jp(j|Y_n)\varphi(j)^Tr_n\bigg)\nonumber\\
&=&r_n+a(n)\bigg(F(r_n,Y_n)-r_n+M_{n+1}(r_n)\bigg).
\end{eqnarray}
where 
\[
F(r,Y)=\varphi(Y)k(Y)+\gamma\varphi(Y)\sum_jp(j|Y)\varphi(j)^Tr-\varphi(Y)\varphi(Y)^Tr+r
\]
and 
\[
M_{n+1}(r)=\gamma\varphi(Y_n)\Big(\varphi(Y_{n+1})^Tr-\sum_jp(j|Y_n)\varphi(j)^Tr\Big).
\]
Define the family of $\sigma$-fields for $n\geq0$:
$$\mathcal{F}_n\coloneqq\sigma(r_0,Y_m, m\leq n).$$
Then $\{M_n(r), \mathcal{F}_n\}$ is a martingale difference sequence for each $r$ satisfying (\ref{MDS_assumption}) for $K_0=2\gamma\|\Phi\|_\infty^2$.  

Since we are working with a time-homogeneous and uncontrolled Markov chain, we can apply the `special case' of our theorem from Remark 1. So we need to show that assumption (\ref{special_contraction}) is satisfied. For ease of notation, we drop subscript `$2$' from $\|r\|_2$ and let $\|r\|$ refer to the Euclidean norm. Let $\langle r,r'\rangle=r^Tr'$ and $\langle x,y'\rangle_D=x^TDy$. Then,
\begin{eqnarray}\label{TD0-complete-1}
    \|\sum_i\pi(i)(F(r,i)-F(s,i))\|^2&=&\|\gamma\sum_i\pi(i)\varphi(i)\sum_jp(j|i)\varphi(j)^T(r-s)\nonumber\\
    &&\;\;\;\;\;\;\;-\sum_i\pi(i)\varphi(i)\varphi(i)^T(r-s) + (r-s)\|^2\nonumber\\
    &=&\|(\gamma\Phi^TDP\Phi-\Phi^TD\Phi+I)(r-s)\|^2\nonumber\\
    &=&\|(\gamma\Phi^TDP\Phi-\Phi^TD\Phi)(r-s)\|^2 \nonumber\\
    &&+(r-s)^T(r-s)\nonumber\\
    &&-2(r-s)^T\Phi^TD\Phi(r-s)\nonumber\\
    &&+(r-s)^T(\gamma\Phi^TDP\Phi+\gamma\Phi^TP^TD\Phi)(r-s).
\end{eqnarray}
Now, 
\begin{eqnarray}\label{TD0-mani1}
(r-s)^T(\gamma\Phi^TDP\Phi+\gamma\Phi^TP^TD\Phi)(r-s)&=&(r-s)^T\gamma\Phi^T(DP+P^TD)\Phi(r-s)\nonumber\\
&=&\gamma \langle \Phi(r-s),P\Phi(r-s)\rangle_D\nonumber\\
&&+\gamma \langle P\Phi(r-s),\Phi(r-s)\rangle_D\nonumber\\
&\stackrel{(a)}{\leq}&2\gamma \|P\Phi(r-s)\|_D\|\Phi(r-s)\|_D\nonumber\\
&\stackrel{(b)}{\leq}&2\gamma\|\Phi(r-s)\|_D^2,
\end{eqnarray}
and 
\begin{eqnarray}\label{TD0-mani2}
2(r-s)^T\Phi^TD\Phi(r-s)&=&2\langle \Phi(r-s), \Phi(r-s)\rangle_D\nonumber\\
&=&2\|\Phi(r-s)\|_D^2.
\end{eqnarray}
Inequality (a) follows from the Cauchy-Schwarz inequality and (b) follows from the fact that $\|Px\|_D\leq\|x\|_D$. Combining (\ref{TD0-mani1}) and (\ref{TD0-mani2}) with (\ref{TD0-complete-1}) gives us:
\begin{eqnarray}\label{TD0-complete-2}
\|\sum_i\pi(i)(F(r,i)-F(s,i))\|^2&\leq&\|r-s\|^2\nonumber\\
&&-2(1-\gamma)\|\Phi(r-s)\|_D^2\nonumber\\
&&+\|(\gamma\Phi^TDP\Phi-\Phi^TD\Phi)(r-s)\|^2.
\end{eqnarray}
 To analyze the last term in (\ref{TD0-complete-2}), we use the fact that the operator norm of a matrix defined as $\|M\|=\sup_{x\neq\theta}\frac{\|Mx\|}{\|x\|}$, using the Euclidean norm for vectors, is equal to the largest singular value of that matrix. Thus
\begin{align}\label{TD0-mani3}
&\|(\gamma\Phi^TDP\Phi-\Phi^TD\Phi)(r-s)\|^2\nonumber\\
&=\|\Phi^T\sqrt{D}(\gamma\sqrt{D}P\Phi-\sqrt{D}\Phi)(r-s)\|^2\nonumber\\
&\leq\lambda_M^2\|(\gamma\sqrt{D}P\Phi-\sqrt{D}\Phi)(r-s)\|^2\nonumber\\
&=\lambda_M^2 \langle(\gamma P-I)\Phi(r-s),(\gamma P-I)\Phi(r-s)\rangle_D\nonumber\\
&=\lambda_M^2 \|(I-\gamma P)\Phi(r-s)\|_D^2\nonumber\\
&\leq \lambda_M^2(1+\gamma)^2\|\Phi(r-s)\|_D^2.
\end{align}
The last inequality follows from the triangle inequality. We now invoke assumption (\ref{TD0_assumption}) and combine (\ref{TD0-mani3}) with (\ref{TD0-complete-2}) as follows:
\begin{eqnarray}\label{TD0-final}
\|\sum_i\pi(i)(F(r,i)-F(s,i))\|^2&\leq&\|r-s\|^2-2(1-\gamma)\|\Phi(r-s)\|_D^2+\lambda_M^2(1+\gamma)^2\|\Phi(r-s)\|_D^2\nonumber\\
&<&\|r-s\|^2-2(1-\gamma)\|\Phi(r-s)\|_D^2\nonumber\\
&&+\left(\frac{\sqrt{2(1-\gamma)}}{1+\gamma}\right)^2(1+\gamma)^2\|\Phi(r-s)\|_D^2\nonumber\\
&=&\|r-s\|^2.
\end{eqnarray}
This gives us the required contraction property with contraction factor $\alpha$ for which an explicit expression can be obtained, using the first inequality in (\ref{TD0-final}), as
\begin{equation*}
        \alpha=\sqrt{1-\min_{r\neq\theta}\frac{\|\Phi r\|_D^2}{\|r\|^2}\bigg(2(1-\gamma)-\lambda_M^2(1+\gamma)^2\bigg)}.
\end{equation*}
Note that as the columns of $\Phi$ are linearly independent, $r\neq\theta\implies\Phi r\neq\theta$ and hence $\frac{\|\Phi r\|_D}{\|r\|}>0$ when $r\neq\theta$. Along with assumption (\ref{TD0_assumption}), this implies that $\alpha<1$.

Let $r^*$ be the fixed point for $\sum_i\pi(i)F(\cdot,i)$, i.e., $\sum_i\pi(i)F(r^*,i)=r^*$. Then,
\begin{align*}
&\sum_i\pi(i)F(r^*,i)=(\Phi^TDk+\gamma\Phi^TDP\Phi-\Phi^TD\Phi+I)r^*=r^* \\
&\implies(\Phi^TDk+\gamma\Phi^TDP\Phi)r^*=\Phi^TD\Phi r^* \\
&\implies (\Phi^TD\Phi)^{-1}(\Phi^TDk+\gamma\Phi^TDP\Phi)r^*=r^* \\
&\implies \Phi(\Phi^TD\Phi)^{-1}\Phi^TD(k+\gamma P\Phi)r^*=\Phi r^* \\
&\implies H(\Phi r^*)=\Phi r^*.
\end{align*}
So iteration (\ref{TD0-modified}) converges to the required fixed point of (\ref{TD0-H-fixed}). Furthermore, assumption (\ref{F_tilde}) also holds with $K=\|k\|_\infty\|\Phi\|_\infty$. The map $r\mapsto F(r,\cdot)$ is clearly Lipschitz. Then Theorem \ref{Main_Theorem} leads to:

\begin{corollary}
Let $n_0\geq N$. Then there exist finite, positive constants $c_1$, $c_2$ and $D$, depending on $\|r_{N}\|$, such that for $\delta>0$ and $n\geq n_0$,  
\begin{enumerate}[(a)]
    \item The inequality
    \begin{equation}
    \|r_n-r^*\|\leq e^{-(1-\alpha)b_{n_0}(n-1)}\|r_{n_0}-r^*\|+\frac{\delta+4a(n_0)c_1}{1-\alpha}
\end{equation}
 holds with probability exceeding \begin{eqnarray}
&&1 \ - \  2M\sum_{m=n_0+1}^n e^{-D\delta^2/\beta_{n_0}(m)}, \ 0 < \delta \leq C, \label{TD_probbound0} \\
&& 1 \ - \  2M\sum_{m=n_0+1}^ne^{-D\delta/\beta_{n_0}(m)}, \ \ \ \delta > C, \label{TD_probbound}
\end{eqnarray}
where $C=e^{\sqrt{M}\left(2\gamma\|\Phi\|_\infty^2\left(1+\|x_{N}\|+\frac{\|k\|_\infty\|\Phi\|_\infty}{1-\alpha}\right)+c_2\right)}$.
\item In particular,
\begin{equation}
\|r_n - r^*\| \leq e^{-(1 - \alpha)b_{n_0}(n-1)}\|r_{n_0} - r^*\| + \frac{\delta+a(n_0)c_1}{1 - \alpha} \ \forall \; n \geq n_0,
\label{TD_mainbound3}
\end{equation}
 with probability exceeding
\begin{eqnarray}
&&1 \ - \  2M\sum_{m \geq n_0+1}e^{-D\delta^2/\beta_{n_0}(m)}, \ 0 < \delta \leq C, \label{TD_probbound1} \\
&& 1 \ - \  2M\sum_{m \geq n_0+1}e^{-D\delta/\beta_{n_0}(m)}, \ \ \ \delta > C. \label{TD_probbound2}
\end{eqnarray}
\end{enumerate}

\end{corollary}

\noindent \textbf{Remark 3:} We mention in passing other  reinforcement learning algorithms where analogous results can be derived, specifically the asynchronous cases of the examples thereof from \citet{Borkar-conc}. The first is the asynchronous version of the Q-learning problem for stochastic shortest path problem \citep{Abounadi} with running cost $k(\cdot,\cdot) \geq 0$, which can be analyzed along the lines of the discounted cost Q-learning above using the fact that the corresponding dynamic programming operator is a contraction w.r.t.\ a weighted max-norm (\citet{BertTsi}, Exercise 3.3, p.\ 325). The asynchronous version of the `post-decision' scheme \citep{Powell} for discounted cost can likewise be covered by the above framework. Note that in case of the above as well as the asynchronous Q-learning scheme for discounted cost studied earlier, it is the asynchrony that puts them beyond the ambit of \citet{Borkar-conc}, necessitating the extension to Markov noise presented here. In case of TD(0), however, Markov noise is already embedded into the scheme itself.

\noindent \textbf{Remark 4:} \textcolor{black}{ Stochastic gradient descent with Lipschitz gradient  can also be thought of as a fixed point seeking scheme for a contraction map, as shown in \cite{Borkar-conc}. Specifically, if the gradient $\nabla f(x)$ is continuously differentiable with a bounded Jacobian (i.e., the Hessian $\nabla^2f(x)$) that is positive definite with its least eigenvalue uniformly bounded away from zero, then for sufficiently small $a > 0$, $-a\nabla f(x) = (x - x\nabla f(x)) - x$, where  the map $x \mapsto x - a\nabla f(x)$ is a contraction w.r.t.\ the euclidean norm. SGD with Markov noise has been studied in \cite{Doan}, \cite{Sun_Tao}, \cite{Wang_Mengdi}, etc. However, our focus here has been in contractive iterates arising in reinforcement learning for approximate dynamic programming.}

\section{Conclusions}

In conclusion, we point out some future directions. Some extensions, e.g., to TD$(\lambda)$ for $\lambda > 0$, may not be very difficult. However, extensions to other cost criteria such as average or risk-sensitive cost are, because their dynamic programming operators are not contractions. Nevertheless, that does not rule out the possibility of building up on these ideas to cover more general ground that will subsume such cases. There are also several other variants of reinforcement learning algorithms left out in this work where even for the discounted cost one might get results in similar spirit, though not of exactly similar form. Finally, such arguments may pave way for regret bounds for reinforcement learning schemes. This needs to be further explored.
\appendix
\section{Appendix A}
Let $\{M_n\}$ be a real valued martingale difference sequence with respect to an increasing family of $\sigma$-fields $\{\mathcal{F}_n\}$. Assume that there exist $\varepsilon, C > 0$ such that
$$E\left[e^{\varepsilon |M_n|}\Big|\mathcal{F}_{n-1}\right] \leq C \ \ \forall \; n \geq 1, \mbox{a.s.}$$
Let $S_n := \sum_{m=1}^n\xi_{m,n}M_m$, where $\xi_{m,n}, \ m \leq n,$, for each $n$, are a.s.\ bounded $\{\mathcal{F}_n\}$-previsible random variables, i.e., $\xi_{m,n}$ is $\mathcal{F}_{m-1}$-measurable $\forall \; m \geq 1$, and $|\xi_{m,n}| \leq A_{m,n}$ a.s.\ for some constant $A_{m,n}$, $\forall \; m, n$. Suppose
$$\sum_{m=1}^nA_{m,n} \leq \gamma_1, \ \max_{1\leq m \leq n}A_{m,n} \leq \gamma_2\omega(n),$$
for some $\gamma_i, \omega(n) > 0, \ i = 1,2; n \geq 1$. Then we have:

\begin{theorem}\label{thm-appendix} There exists a constant $D > 0$ depending on $\varepsilon, C, \gamma_1, \gamma_2$ such that for $\epsilon > 0$,
\begin{eqnarray}
P\left(|S_n| > \epsilon\right) &\leq& 2e^{-\frac{D\epsilon^2}{\omega(n)}}, \ \ \mbox{if} \ \epsilon \in \left(0, \frac{C\gamma_1}{\varepsilon}\right], \label{LW1} \\
&&  2e^{-\frac{D\epsilon}{\omega(n)}},  \ \ \mbox{otherwise.} \label{LW2}
\end{eqnarray}
\end{theorem}

This is a variant of Theorem 1.1 of \citet{LW}. See \citet{Gugan}, Theorem A.1, pp.\ 21-23, for details.

\section{Appendix B: \textcolor{black}{Lipschitz Constants}}
\textcolor{black}{\\
\noindent \textbf{\normalsize Part (i) - Stationary Distribution} \\
\noindent We first give some bounds for the Lipschitz constant of the map $x\mapsto\pi_x$ where $x\in\mathcal{R}^d$ and $\pi_x$ is the stationary distribution corresponding to the transition probabilities $p_x(\cdot|\cdot)$. Using section 3 of \citet{Meyer}, we have
\begin{eqnarray*}
\sum_{i\in S}|\pi_x(i)-\pi_y(i)|=\|\pi_x-\pi_y\|_1&\leq&\kappa_i \|P_x-P_y\|_\infty \\
&\leq& \kappa_i L_1\|x-y\|
\end{eqnarray*}
Here $\|P_x-P_y\|_\infty$ denotes the operator norm of matrix $(P_x-P_y)$ under the $\ell_\infty$ norm for vectors and is equal to the largest $\ell_1$ norm of the rows of $(P_x-P_y)$. $\kappa_i$ denotes one of the condition numbers of the Markov chain as defined in \citet{Meyer}. For our case $\kappa_i$ can be $\kappa_1$, $\kappa_2$, $\kappa_5$, or $\kappa_6$, out of which the smallest is $\kappa_6$, defined using the ergodicity coefficient as defined in \citet{Seneta}. So, $L_2=L_1\kappa_i$. Alternatively, if we assume continuous differentiability of the map $w \mapsto p_w(\cdot|\cdot)$, then the explicit formula for gradient of $\pi_w$ is a special case of the formula in Proposition 1 of \citet{Marbach} (see also \citet{Lasserre}). These can be used to bound the Lipschitz constant. In fact, since Lipschitz constant does not change on convolution with a smooth probability density, one can use the aforementioned results to get a Lipschitz constant via smooth approximations.\\
}

\noindent \textbf{\normalsize Part (ii) - Solution of Poisson Equation} \\

\noindent We define $V(\cdot,\cdot) = [V^1(\cdot,\cdot) : V^2(\cdot,\cdot) : \cdots\cdots : V^d(\cdot, \cdot)]^T$ to be a solution of the Poisson equation:
\begin{equation*}
    V(x,i)=F(x,i)-\sum_{j\in \mathcal{S}}\pi_x(j)F(x,j)+\sum_{j\in \mathcal{S}}p_x(j|i)V(x,j), \ i\in S.
\end{equation*}
For each $x$ and $\ell$,  the Poisson equation specifies $V^\ell(x,\cdot)$ uniquely only up to an additive constant. Adding or subtracting a scalar $c^\ell(x)$ to $V^\ell(x,i)$ for each state $i$ still gives us a solution of the Poisson equation. So we add the additional constraint that $V(x,i_0)=0,\ \forall x$ for some prescribed $i_0\in S$. With this additional constraint, the system of equations given by (\ref{Poisson_eqn}) has a unique solution. Thus let $V$ denote the unique solution of the set of equations parametrized by $x$, given by $V(x,i_0)=0 \ \forall x$ and
$$V(x,i)=F(x,i)-\sum_{j\in \mathcal{S}}\pi_x(j)F(x,j)+\sum_{j\in \mathcal{S}}p_x(j|i)V(x,j), \ i\in S\setminus\{i_0\}.$$
We next show that the mapping $x\mapsto V(x,i)$ is Lipschitz for all $i\in S$. Note that for $i\in S\setminus\{i_0\}$,
\begin{eqnarray*}
V(x,i)-V(z,i)&=& F(x,i)-F(z,i) \\
&&+ \sum_{j\in \mathcal{S}}\pi_x(j)F(x,j)-\sum_{j\in \mathcal{S}}\pi_z(j)F(z,j) \\
&&+ \sum_{j\in \mathcal{S}}p_x(j|i)V(x,j) - \sum_{j\in \mathcal{S}}p_z(j|i)V(z,j) \\
&=& F(x,i)-F(z,i)\\
&&+ \sum_{j\in \mathcal{S}}\pi_x(j)\left(F(x,j)-F(z,j)\right) + \sum_{j\in \mathcal{S}}\left(\pi_x(j)-\pi_z(j)\right)F(z,j) \\
&&+ \sum_{j\in \mathcal{S}}p_x(j|i)\left(V(x,j)-V(z,j)\right) + \sum_{j\in \mathcal{S}}\left(p_x(j|i)-p_z(j|i)\right)V(z,j) \\
\end{eqnarray*}
Note that $\sum_{j\in \mathcal{S}}p_x(j|i)\left(V(x,j)-V(z,j)\right)=\sum_{j\in \mathcal{S}\setminus\{i_0\}}p_x(j|i)\left(V(x,j)-V(z,j)\right)$. The above are then a set of $(|S|-1)$ equations, where $|S|$ is the size of the finite state space, with variables $V(x,i)$ and $V(z,i)$ for $i\in S\setminus\{i_0\}$. Each variable is itself a vector in $\mathcal{R}^d$ but the $d$ components of the variables are independent in the above set of equations. So, we can work with each component of the above set of equations separately. Let $P_x^{-\{i_0\}}$ be the sub-stochastic matrix obtained by removing the row and column corresponding to $i_0$ from the transition matrix $P_x$ of the Markov chain. Then the $\ell^{\textrm{th}}$ component of the above set of $(|S|-1)$ equations can be represented as
\begin{eqnarray*}
\left(I-P_x^{-\{i_0\}}\right)\left(V^\ell(x)-V^\ell(z)\right)&=&(F^\ell(x)-F^\ell(z))+\Lambda_x(F^\ell(x)-F^\ell(z))\\
&&+(\Lambda_x-\Lambda_z)F^\ell(z)+\left(P_x^{-\{i_0\}}-P_z^{-\{i_0\}}\right)V^\ell(z).
\end{eqnarray*}
Here $V^\ell(x)$ and $F^\ell(x)$ are vectors in $\mathcal{R}^{|S|-1}$ containing values $V^\ell(x,i)$ and $F^\ell(x,i)$, respectively, for all states $i\in S\setminus\{i_0\}$ and $I$ is the identity matrix of dimension $|S|-1$. For example, $V^\ell(x)=[V^\ell(x,1),\cdots, V^\ell(x,i_0-1),V^\ell(x,i_0+1),\cdots,V^\ell(x,|S|)]^T$. $\Lambda_x$ is the $(|S|-1)\times(|S|-1)$ matrix with identical rows and each row as the vector with values $\pi_x(i)$ for $i\in S\setminus\{i_0\}$. As explained before, this set of equations have a unique solution and hence the matrix $\left(I-P_x^{-\{i_0\}}\right)$ is invertible. Using this, we get 
\begin{eqnarray*}
\|V^\ell(x)-V^\ell(z)\|_\infty&=&\|\Big(I-P_x^{-\{i_0\}}\Big)^{-1}\Big((F^\ell(x)-F^\ell(z))+\Lambda_x(F^\ell(x)-F^\ell(z))\\
&&\;\;\;\;+(\Lambda_x-\Lambda_z)F^\ell(z)+\left(P_x^{-\{i_0\}}-P_z^{-\{i_0\}}\right)V^\ell(z)\Big)\|_\infty\\
&\leq& \|(I-P_x^{-\{i_0\}})^{-1}\|_\infty\Big(2L_3\|x-z\|\\
&&\;\;\;\;+\|\Lambda_x-\Lambda_z\|_\infty\|F^{\ell}(z)\|_\infty+\|P_x^{-\{i_0\}}-P_z^{-\{i_0\}}\|_\infty\|V^{\ell}(z)\|_\infty\Big)\\
&\leq& \|(I-P_x^{-\{i_0\}})^{-1}\|_\infty\Big(2L_3+L_2\|F^{\ell}(z)\|_\infty+L_1\|V^{\ell}(z)\|_\infty\Big)\|x-z\|.
\end{eqnarray*}
Here for matrices $\|\cdot\|_\infty$ denotes the operator norm of the matrix under the $\ell_\infty$ norm for vectors. Note that $\|\Lambda_x-\Lambda_z\|_\infty=\sum_{i\in S\setminus\{i_0\}}|\pi_x(i)-\pi_z(i)|$. 

Since
$$(I-P_x^{-\{i_0\}})^{-1}  = \sum_{m=0}^\infty (P_x^{-\{i_0\}})^m,$$
the $(i,j)$\textsuperscript{th} element of this matrix is the expected number of visits to $j$ starting from $i$ before hitting $i_0$. Summing over $j$, we get the expected hitting time of $i_0$ starting from $i$, which serves as a componentwise bound on this matrix. There are standard bounds on the mean hitting times in terms of a suitable stochastic Liapunov function \citep{Meyn_book}. 
By Lemma \ref{Bound_x_n}, we know that $x$ lies in a compact domain and hence $\|F^\ell(x)\|_\infty$ and $\|V^\ell(x)\|_\infty$ are bounded. By bounding the $\ell_\infty$ norm of $(V^\ell(x)-V^\ell(z))$, we have shown that $x\mapsto V^\ell(x,i)$ is Lipschitz for all $i\in S$, $\ell\in\{1,\cdots,d\}$ and $\|x\|\leq x_{n_0}+\frac{K}{1-\alpha}$ with common Lipschitz constant $L$. As discussed, its magnitude will depend on mean hitting times. We do not get into the details here.

\end{document}